\documentclass[10pt,twocolumn,letterpaper]{article}

\usepackage{cvpr}
\usepackage{times}

\usepackage{amsmath}
\usepackage{amssymb}

\usepackage{epsfig}
\usepackage{graphicx}
\usepackage{color}

\usepackage[T1]{fontenc}
\usepackage[utf8]{inputenc}

\usepackage{booktabs}
\usepackage{rotating}
\usepackage{tabularx}
\usepackage{arydshln}

\usepackage{pifont}

\usepackage{siunitx}

\newcommand\floor[1]{\lfloor#1\rfloor}

\newenvironment{Turn}{\begin{turn}{90}\begin{minipage}{1.6cm}\raggedright}
{\end{minipage}\end{turn}}

\usepackage[pagebackref=true,breaklinks=true,letterpaper=true,colorlinks,bookmarks=false]{hyperref}

\cvprfinalcopy 

\def\cvprPaperID{18} 

\ifcvprfinal\fi
\begin{document}

\title{Spatial Sampling Network for Fast Scene Understanding}

\author{Davide Mazzini \qquad Raimondo Schettini \\
University of Milano-Bicocca\\
{\tt\small \{davide.mazzini, raimondo.schettini\}@unimib.it}
}

\maketitle

\begin{abstract}
We propose a network architecture to perform efficient scene understanding. This work presents three main novelties: the first is an Improved Guided Upsampling Module that can replace \emph{in toto} the decoder part in common semantic segmentation networks.
Our second contribution is the introduction of a new module based on spatial sampling to perform Instance Segmentation. It provides a very fast instance segmentation, needing only a thresholding as post-processing step at inference time. Finally, we propose a novel efficient network design that includes the new modules and test it against different datasets for outdoor scene understanding. To our knowledge,  our network is one of the most efficient architectures for scene understanding published to date, furthermore being 8.6\% more accurate than the fastest competitor on semantic segmentation and almost five times faster than the most efficient network for instance segmentation.
\end{abstract}

\section{Introduction}
Most of the current architectures that perform semantic segmentation rely on an encoder-decoder architecture. This is the simplest and most effective way to design the model in order to increase the receptive field of the network at the same time keeping the computational cost feasible. To mitigate the effect of information loss caused by downsampling, state-of-the-art architectures make use of additional ways to increase the receptive field, e.g. dilated convolutions \cite{drn}, ASPP \cite{deeplabv3}, PSP \cite{psp}. However the use of dilated convolutions results in computational heavy architectures and, in most cases, the best trade-off consists of a mix of downsampling and dilation. As a consequence, common network architectures, employ upsampling operators to output a semantic map with the same resolution as the input.

Our \textbf{first contribution} consists of a new module named Improved Guided Upsampling Module i.e. iGUM. It replaces traditional upsampling operators like bilinear and nearest neighbor and can be plugged into any existing architecture and trained end-to-end within the network. With a low additional computational cost, it helps to improve the prediction along object boundaries when upsampling output probability maps of semantic segmentation networks. We discovered that it can replace the entire decoder part of efficient semantic segmentation networks, obtaining a consistent speed-up and even an improvement of performance in most cases.

As our \textbf{second contribution} we designed a novel module that model Semantic Instance Segmentation as a diffusion process through a differentiable sampling operator. The resulting layer has three main advantages: It is computationally lightweight allowing for very fast inference at test time. It can be trained end-to-end with the whole network and requires only a thresholding as post-processing step.

\begin{figure}[t]
\centering
\includegraphics[width=\linewidth]{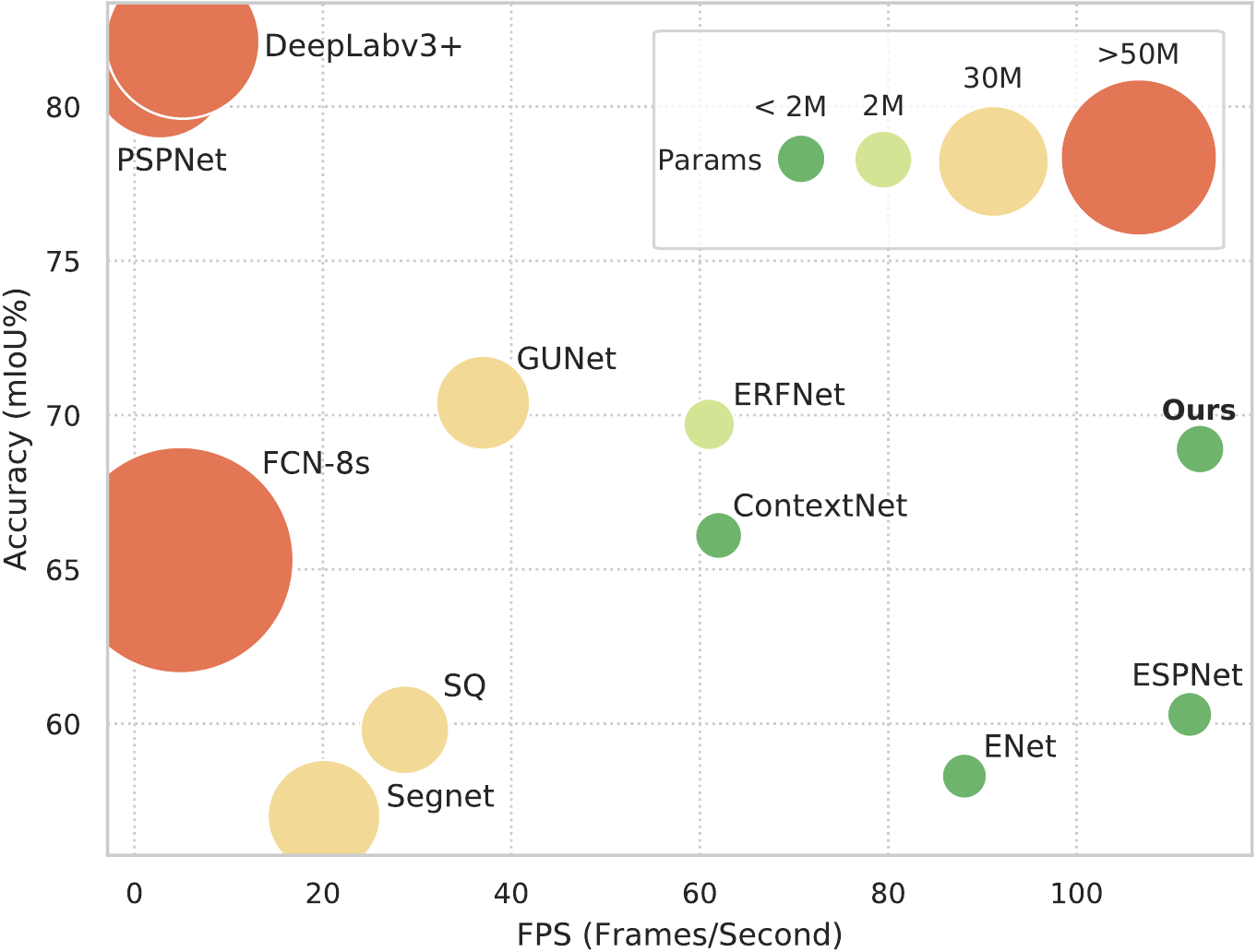}
\caption{Speed vs mIoU computed on Cityscapes test set for semantic segmentation architectures. Points are proportionally sized with respect to the number of parameters of the model.}
\label{fig:miou_vs_fps}
\end{figure}

\begin{figure*}[ht]
\centering
\includegraphics[width=0.75\linewidth]{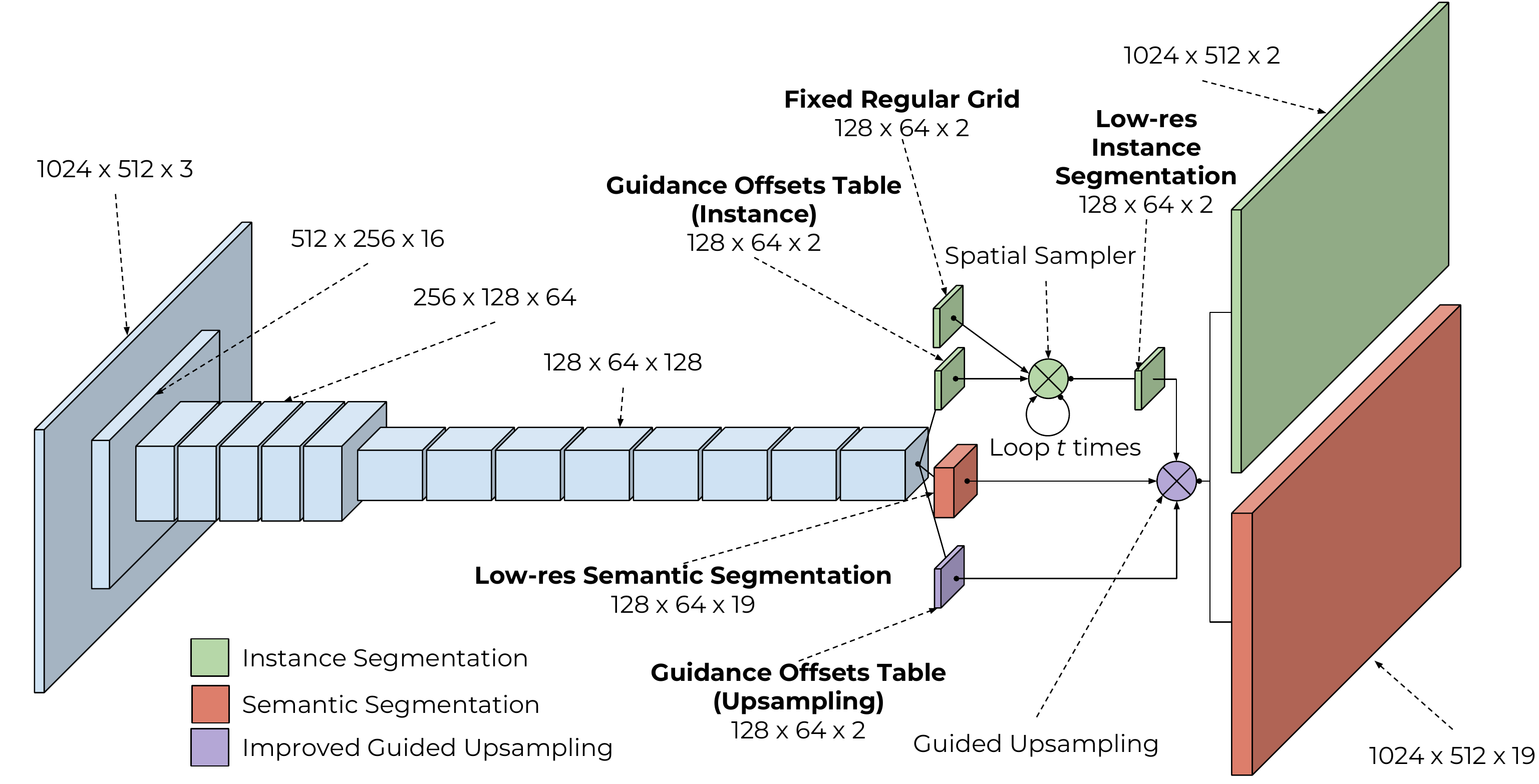}
\caption{Our network design. The decoder is composed of three parts: the semantic segmentation part, the Instance Segmentation module and the Improved Guided Upsampling Module}
\label{fig:network_design}
\end{figure*}

Our \textbf{third contribution} consists of a novel lightweight neural architecture to perform scene understanding in applications where speed is a mandatory requirement.
We run experiments to assess different aspects of our modules and our complete architecture on three different datasets: Camvid \cite{camvid}, PASCAL VOC 2012 \cite{pascalvoc} and Cityscapes \cite{cityscapes}. Our network is able to achieve 68.9\% of mIoU on the popular Cityscapes dataset at 113 FPS on a Titan Xp GPU being 8.6\% more accurate than the fastest published network up to date.

\section{Related Works}
\label{sec:related_works}

\textbf{Efficiency-oriented architectures}
SegNet\cite{segnet}, one of the first efficiency-oriented architectures together with \cite{sq}, introduced an efficient way to exploit high-resolution information by saving max-pooling indices from the encoder and using them during upsampling.
ENet authors \cite{enet} designed the first highly-efficient architecture by making use of clever design patterns and state-of-the-art building blocks: residual blocks \cite{resnet}, early downsampling, and 1D factorized convolutions \cite{inceptionv1}. We think that the use of factorized convolutions is the main reason for the success of this architecture. ERFNet \cite{erfnet} authors implemented an architecture with a very similar overall structure to ENet but with a residual module named Non-Bottleneck-1D module. The overall architecture raises the performance of ENet of a great margin. We will build our fast architecture on top of ERFNet. ESPNet \cite{espnet} is a very recent, efficient architecture that makes use of a novel module named ESP module. It achieves a very good tradeoff in terms of accuracy and network speed.

\textbf{Instance Segmentation architectures}
Following \cite{discriminativeloss} we categorize instance segmentation architectures in four different classes: Proposal-based, Recurrent methods, Energy-based and Clustering.
\cite{hariharan2014simultaneous,chen2015multi} are examples of proposal-based approaches. They employ MCG \cite{mcg} as a class-agnostic predictor of object proposals and they subsequently make use of a classification step to produce instances.
A good number of recent works rely on the joint use of an object detector and a semantic segmentation network \cite{dai2015convolutional, dai2016instance, arnab2016bottom, hayder2016shape}. As a matter of fact, recent state-of-the-art instance segmentation algorithms \cite{panet,maskrcnn} are based on enhanced versions of the Faster R-CNN object detector \cite{fasterrcnn} but, from the point-of-view of this work, architectures like PANet \cite{panet} or Mask-RCNN \cite{maskrcnn}, are still computationally too heavy to be employed on edge devices for real-time applications.

Other works like \cite{stewart2016end} fall into the category of recurrent methods, i.e. they adopt recurrent networks to generate instances in a sequential way. \cite{romera2016recurrent} uses LSTMs to output binary segmentation maps for each instance. \cite{ren1605end} enhance the approach of \cite{romera2016recurrent} by adding an object detector network to refine the output.
\cite{bai2017deep, kirillov2017instancecut} uses alternative ways to detect instances. \cite{bai2017deep} exploit the watershed transform. \cite{kirillov2017instancecut} uses an approach based on CRF together with a customized MultiCut algorithm. The last category of methods involves the transformation of the input image into a representation that is afterward clustered into a set of discrete instances \cite{silberman2014instance,zhang2015monocular}.
 \cite{discriminativeloss} introduces a novel loss function that induces an embedding space representing separate instances. As a second step, it employs a clustering algorithm to extract the segmented instances. Instead \cite{uhrig2016pixel,kendall2017multi} trains a network to predict a dense vector field where each vector points towards the instance center. The fastest architectures for instance segmentation up to date belong to this category, see also Table \ref{tab:instance_sota}. However, accurate clustering algorithms are quite heavy to run on a large set of points (pixels) with high-resolution images. This makes impractical to use this category of algorithms in real settings with real-time processing pipelines.

Our network architecture performs instance segmentation in a similar way to the clustering methods. However, it makes the use of a clustering algorithm unnecessary by exploiting a very simple and efficient iterative sampling algorithm. We will explain in details our method for instance segmentation and outline the differences with \cite{uhrig2016pixel,kendall2017multi} in Section \ref{sec:instance}.

\section{Guided Upsampling Module}
\label{sec:gum}
Most state-of-the-art semantic segmentation networks are based on encoder-decoder architectures \cite{deeplabv3plus, psp, deeplabv3, mapillary, resnet38, fcn8s, enet, erfnet,espnet}. Since the task involves a dense per-pixel prediction, usually a probability vector over the classes distribution is predicted for each pixel. This is arguably inefficient both computationally and from a memory occupation point-of-view.
A more efficient output representation would involve a non-uniform density grid which follows the objects density distribution in the scene. A step towards this direction has been presented in \cite{gum} with the introduction of the Guided Upsampling Module (GUM). GUM is an upsampling operator built to efficiently handle segmentation maps and improve them along objects' boundaries. Classical upsampling operators (e.g. nearest neighbor or bilinear) make use of a regular grid to sample from the low-resolution image. GUM introduces a warping grid named \emph{Guidance Offset Table} to correct the prediction map along object boundaries. The Guidance Offset Table is predicted by a neural network branch named \emph{Guidance Module}. The whole module can be plugged-in in any existing architecture and trained end-to-end. Given $V_i$ the output feature map and $U_{nm}$ the input feature map, GUM over the nearest neighbor operator can be defined as follows:
\begin{equation} \label{eq:nearestguided}
  \small
\begin{split}
  V_i = \sum^H_n{\sum^W_m{U_{nm}(\delta(\floor{x^s_i + p_i + 0.5} - m)}}\\
  \delta(\floor{y^s_i + q_i + 0.5} - n))
\end{split}
\end{equation}
where $x^s_i$ and $y^s_i$ are the spatial sampling coordinates. $\floor{x^s_i + 0.5}$ rounds coordinates to the nearest integer location and $\delta$ is a Kronecker function. $p_i$ and $q_i$ are what makes GUM different from nearest-neighbor: two offsets that shifts the sampling coordinates of each grid element in $x$ and $y$ dimensions respectively. They are the output of a function $\phi_i$ of $i$, the Guidance Module, defined as: $\phi_i = (p_i, q_i)$.
For the definition of GUM over bilinear sampling refer to \cite{gum}.
The resulting operator is differentiable with respect to $U$, $p_i$ and $q_i$. In \cite{gum}, the GUM module is applied to the network output probability map, even though it could be employed anywhere within the architecture.
\subsection{Improved Guided Upsampling Module}\label{sec:igum}
\begin{figure}
\centering
\includegraphics[width=\linewidth]{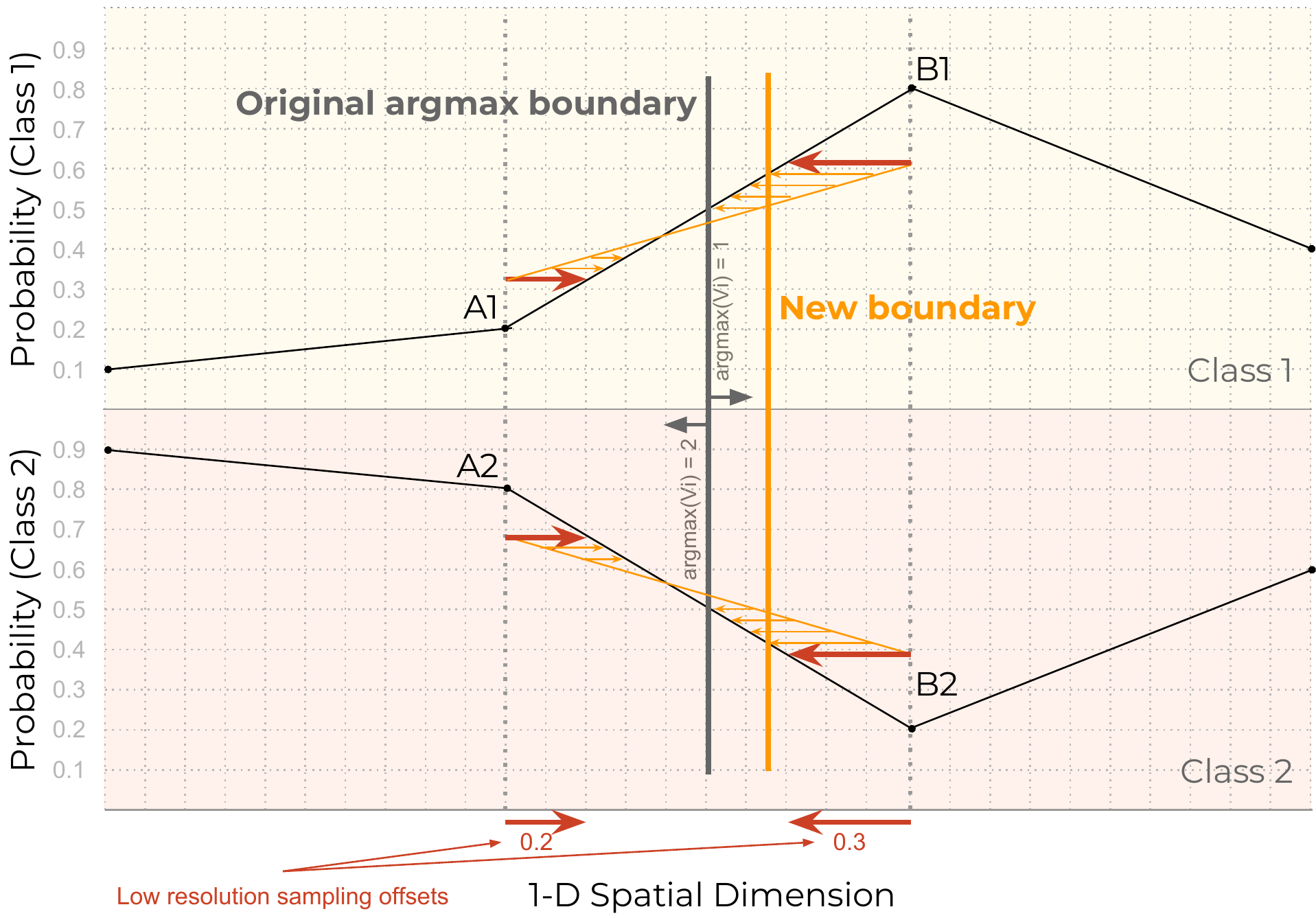}
\caption{ A toy example: upsampling a 1D probability map with two pixels ($A$ and $B$ on the horizontal axis) and two classes ($1$ and $2$ on the vertical axis). The argmax boundary can be moved in two ways: with the orange arrows i.e. GUM \cite{gum}, or with the red arrows i.e. Improved GUM. The latter formulation requires only two sampling offsets.}
\label{fig:igum_boundaries}
\end{figure}
\begin{figure*}[ht]
\centering
\includegraphics[width=0.70\linewidth]{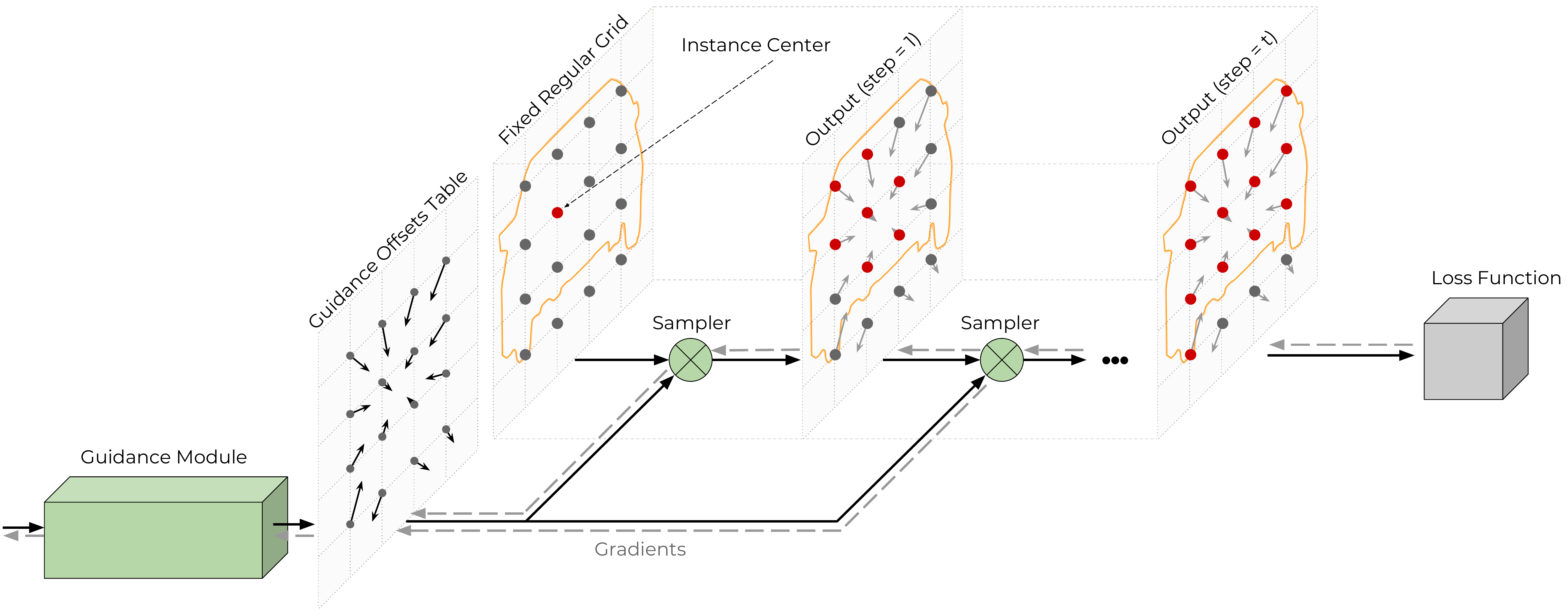}
\caption{Instance Segmentation Module: a Guidance Offsets Table is used to iteratively sample values from the instance center and spread it to all the neighboring pixels.}
\label{fig:instance_segmentation}
\end{figure*}
Let $U_{cnm} \in \mathbb{R}^{C\times N\times M}$ be an output probability map to be upsampled, where $C$ represents the number of classes, $N$ and $M$ are the spatial dimensions. The GUM module needs a \emph{Guidance Offsets Table} of the same spatial dimensionality as the output features map. Let the upsampling factor be $f$. To produce a feature map $V_{cnm} \in \mathbb{R}^{C\times fN\times fM}$ the GUM needs to predict a guidance offsets table of cardinality $2 \times fN\times fM $. This is certainly more efficient than predicting the full resolution probability map, especially if $C$
is large because, regardless of the cardinality of classes, the network has to predict only a 2-dimensional vector (offset) for each pixel in the output map. However, a lightweight decoder structure is still needed. Our desiderata are to reduce even more the computational burden and to completely remove the decoder part of the network.

\textbf{A geometric view on GUM} Figure \ref{fig:igum_boundaries} depicts a toy example on the use of GUM to upsample an output probability map. The spatial dimension is visualized on the x-axis. In this example, only one dimension is represented although in the real problem there are two spatial dimensions. On the y-axis is represented the probability over the two classes. The two dashed vertical lines indicate two points in the low-resolution probability map. Pairs of black points lying on these lines, i.e ($A1$, $A2$) and ($B1$, $B2$) represent the predicted probability vector over the two classes ($1$ and $2$). The continuous black lines represent the sub-pixel values of the probability distribution obtained by linear interpolation. Notice that, this visualization in a continuous space, allow us to abstract from the upsampling factor. A gray vertical line is depicted on the \emph{argmax} boundary. i.e. the point where the probability distribution is equal over the two classes and where the \emph{argmax} value changes.
By looking at Figure \ref{fig:igum_boundaries} it is clear that the position of the original boundary is dependent on the values of the four points $A1$, $A2$, $B1$, $B2$. Thus, the boundary subpixel position is tied to the two adjacent probability vectors. The idea of GUM is represented in Figure \ref{fig:igum_boundaries} by the multiple thin parallel orange arrows. They are spatial sampling vectors predicted by the network at high resolution. The probability value is sampled, i.e. copied, from the head to the tail of each arrow. With this sampling operation, it is possible to move the \emph{argmax} boundary as shown in Figure \ref{fig:igum_boundaries}. Notice that the orange parallel arrows are the same for the two classes: i.e. the boundary can be moved by predicting the same arrows for every class.

\textbf{Improved GUM intuition} Our intuition is that the argmax boundary can be moved by predicting only the low-resolution sampling offsets, i.e. the red arrows at the bottom of Figure \ref{fig:igum_boundaries}. (The other bold red arrows are the same arrows moved along the vertical axis). By looking at the parallel thin arrows in between, we observe that they can be obtained by interpolating the bold arrows. We can move the \emph{argmax} boundary by predicting only one additional value low-resolution pixel. Intuitively we can extend this concept to two dimensions by predicting a 2D spatial offsets vector for each spatial location and obtain the other sampling offsets by bilinear interpolation. To summarize, a network with the Improved Guided Upsampling Module, results in a very simple structure. It is composed by an encoder with two branches: the first predicts the output probability map and the second predicts a low-resolution \emph{Guidance Offsets Table}. Both have spatial dimension $N\times M$. The Guidance Offsets Table is then upsampled to the target resolution $fN\times fM$ depending on the upsampling factor $f$. Finally the high resolution \emph{Guidance Offsets Table} is given as input to the GUM module as in \cite{gum}. The resulting module can be plugged into any common CNN architecture and trained end-to-end with the whole network.

\begin{figure}
\centering
\includegraphics[width=0.45\linewidth]{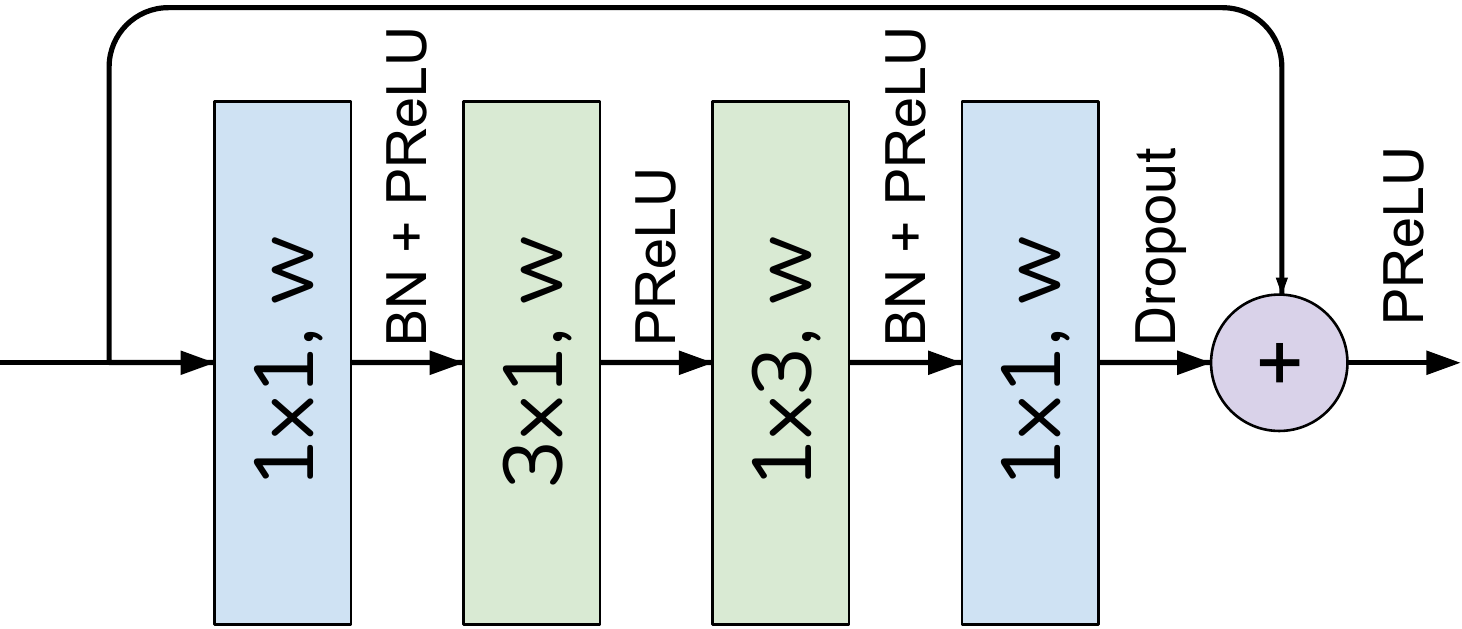}
\caption{Our proposed Lightweight Non-Bottleneck Module. $w$ represents the number of input channels. $d_1 \times d_2, f$ represents the kernel sizes $(d_1,d_2)$ and the numbere of output channels $f$.}
\label{fig:lightweight_nbt_1d_module}
\end{figure}
\begin{figure}
\centering
\includegraphics[width=0.7\linewidth]{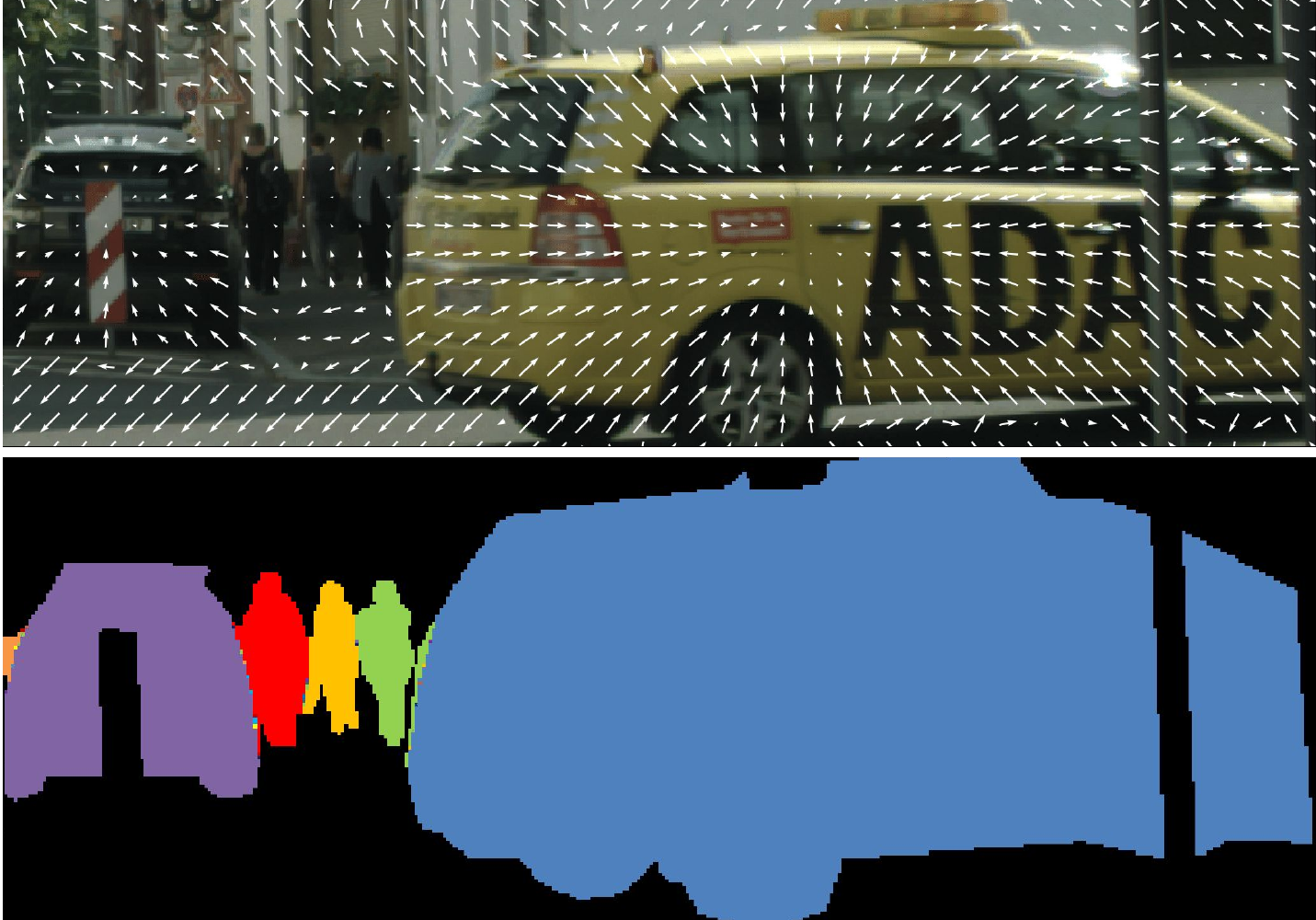}
\caption{Top: visualization of the Guidance Offsets Table of the Instance Module. Bottom: visualization obtained by coloring with a different color each unique value in the output map.}
\label{fig:instance_visualization}
\end{figure}
\section{Instance Segmentation Module}
\label{sec:instance}
The most efficient architectures for instance segmentation e.g. \cite{kendall2017multi,neven2017fast} (see Section \ref{sec:related_works}) are trained to produce a dense output map with a particular form of embedding for each pixel. To extract every single instance, the embedding needs to be post-processed by a clustering algorithm. Usually, the time needed to run the clustering algorithm on the raw output is not considered when evaluating the speed of state-of-the-art methods. As a matter of fact, running a clustering algorithm on a high-resolution output is computationally intensive, making these approaches inefficient in real-world scenarios. Consider for example the mean-shift algorithm \cite{meanshift} employed by \cite{discriminativeloss}. It has a complexity of $O(Tn^2)$ where $T$ is the number of iterations and $n$ is the number of points in the data set. In case of Cityscapes dataset where the resolution of the images is $1024 \times 2048$ the number of points is 2 Million and the number of operations can easily scale up to 20 G-FLOPs, which is roughly the amount of operations needed to perform a forward pass on a VGG-19 architecture \cite{vgg} (one of the heaviest architectures to date \cite{gigi}).

\textbf{Instance Segmentation by Sampling} We propose an instance segmentation module that associates to each pixel the instance centroid, similarly to \cite{kendall2017multi,neven2017fast}. We train our network with an  $L^2$ regression loss applied directly to the network output:
\begin{equation} \label{eq:instance_loss}
    \mathcal{L_{\text{Instance}}} = \frac{1}{N}\sum_{n \in N}\lVert i_n - \hat{i}_n \lVert_2
\end{equation}
where $i \in \mathbb{R}^2$ is the instance center in pixel coordinates. $\hat{i}_n \in \mathbb{R}^2$ is the vector predicted by the network. The loss is averaged over every labeled pixel $n \in N$ in a minibatch. The peculiar trait of our method lies in the architecture. Figure \ref{fig:instance_segmentation} shows how the Instance Segmentation module works: a \emph{Guidance Module} predicts a \emph{Guidance Offsets Table} of vectors (bearing the terminology from \cite{gum}). The usual way to train instance segmentation networks \cite{kendall2017multi} is to directly apply the loss function in Eq.\ref{eq:instance_loss} to the Guidance Offsets Table. In our architecture instead, a differentiable sampler, like the one used in \cite{gum} or \cite{stn}, samples 2D points from a fixed regular grid using the Guidance Offsets Table. The fixed regular grid codifies the 2D coordinates of the exact location of each pixel. This sampling process is applied $t$ times \emph{sharing the same} Guidance Offsets Table (see Figure \ref{fig:instance_segmentation}). Thus, by consecutive sampling steps, the 2D coordinate values of instance centers are spread all over the area covered by the instance. Finally, the loss function in Eq. \ref{eq:instance_loss} is computed on the sampled output. Figure \ref{fig:instance_segmentation} shows the gradient flow: notice that, it flows backward through every sampler step but not to the fixed regular grid. The only way to decrease the loss function is to produce a Guidance Offsets Table with vectors that point towards the instance center.

There is a main advantage of this approach over the classical method of training directly the vector field: in an ideal setting, if the network predicts a perfect output, all the vectors associated with a particular instance will points precisely to the instance center. This doesn't happen in practical cases, thus, the need of a clustering algorithm. Here emerges the major advantage of our approach: by sampling multiple times, a diffusion process is generated. As a consequence, the value of the instance center propagates through all the pixels associated with that particular instance. If a far vector points to an imprecise location towards the instance center, by successive sampling steps, the center value propagates to increasing areas eventually covering all the instance area.
Top visualization in Figure \ref{fig:instance_visualization} shows the Guidance Offsets Table overimposed on the input image. Note that the vectors' magnitude decrease drastically near the instance center. The visualization at the bottom of Figure \ref{fig:instance_visualization} is obtained by summing for each pixel location the 2D centroids' predictions coordinates. A single value per location is obtained (it can be interpreted as a unique instance identifier). In Figure \ref{fig:instance_visualization} a different color is assigned to \emph{each} unique value.
The only post-processing step applied to the final output is a thresholding over the instance area. Very small false positives are eliminated by this step, they can be noticed by zooming Figure \ref{fig:instance_visualization} (bottom) along object boundaries.

\textbf{Two Practical Tips} First, to simplify the learning process, the values of the Guidance Offsets Tables both for the Improved GUM and for the Instance Module are limited to the interval $[-1,1]$ by mean of a $tanh(x) = \frac{e^x - e^{-x}}{e^x + e^{-x}}$ function. Second, the GUM operator used to upsample the instance output is based on Bilinear sampling at training time, and Nearest-neighbor at test time. Nearest-neighbor is the right choice to upsample unique id values but it is not directly differentiable w.r.t the Guidance Offsets Table.

\section{Network Architecture}
\label{sec:architecture}
Our architecture for semantic segmentation consists of a lightweight encoder and an iGUM layer as decoder. For the task of instance segmentation we added the Instance Segmentation Module before the guided upsampling (refer to Figure \ref{fig:network_design} for details). The iGUM is described in details in Section \ref{sec:gum} whereas the encoder has a very simple structure inspired by \cite{erfnet}. However, the main building block is a novel Lightweight Non-bottleneck Module described in this section.

\textbf{Encoder-only architecture} In encoder-decoder architectures, the decoder plays a refinement role where features are subsequently upsampled to match the input size and to finally output the dense  per-pixel prediction. By employing our iGUM module we are able to completely remove the decoder part: as shown in Table \ref{tab:gum_nogum} this a major factor for a highly-efficient semantic segmentation network.

\textbf{Lightweight Non-Bottleneck-1D Module} The Non-Bottleneck-1D module has been proposed by Romera et al. as the main building block of ERFNet \cite{erfnet}. It is a residual block composed of 1D convolutions also known as \emph{Asymmetric convolutions} \cite{edanet}. Numerous works investigated decomposed filters from a theoretical point of view, e.g. \cite{sironi2015learning, alvarez2016decomposeme}) and in practical settings , e.g. \cite{inceptionv3, inceptionv4, enet, edanet}. The idea is that each convolutional layer can be decomposed in 1D filters that can additionally include a non-linearity in between.
The decomposed layers have intrinsically lower computational cost than the corresponding full rank counterparts. Our Lightweight Non-Bottleneck Module consists in a very simple design: it is a non-bottleneck residual block with two asymmetric kernels $(3\times1)$ and $(1\times3)$ preceded and followed by $1\times1$ channelwise convolutions. The module design is depicted in Figure \ref{fig:lightweight_nbt_1d_module}. We discovered this module to be particularly effective as part of our architecture. We motivate every design choice experimentally within the ablation study in Section \ref{sec:cityscapes_results}.

\textbf{Early downsampling}
Following the last works on efficient models \cite{enet,drn,erfnet,fastdownsampling} our architecture employs an early downsampling strategy to speed-up inference time.
In our network, the first layers act as early feature extractors whereas the most intense operations are carried out by the inner network modules to favor a more complex representation in late activations.
Our downsampling block, inspired by \cite{erfnet, enet}, performs downsampling by concatenating the parallel outputs of a single $3\times3$ convolution with stride 2 and a Max-Pooling module.
\section{Experiments}\label{sec:experiments}
We built our architecture based on different types of experiments to verify each design hypothesis. After presenting the implementation details for the sake of reproducibility, we introduce different experiments concerning the Improved Guided Upsampling Module and in-depth ablation studies on our architecture. Finally, we compare our network with state-of-the-art efficient architectures on different datasets for semantic and instance segmentation.

\textbf{Evaluation metrics} \emph{mean of class-wise Intersection over Union (mIoU)} is used to measure semantic segmentation quality on all datasets. It is computed as the classwise mean of the intersection over union measure. On Camvid dataset also the class average and the global average are computed being the mean of the accuracy on all classes and the global pixel accuracy respectively. \emph{Frame Per Second (FPS)} is used as speed measure, defined as the inverse of time needed for our network to perform a single forward pass. FPS have been computed on a single Titan Xp GPU whether not differently specified. Following \cite{enet}, we removed all Batch-Normalizations at test time merging them with close convolutions.

\textbf{Training Recipes}
All experiments have been conducted within the Pytorch framework \cite{pytorch} v1.0. For training, following \cite{erfnet} we use the Adam optimizer \cite{adam} in an initial learning rate of $5e^{-4}$ and weight decay of $1e^{-4}$. The learning rate is scheduled by multiplying the initial learning rate by $(1 - \frac{epoc}{maxEpochs})^{0.9}$. All models are trained for 150 epochs with a mini-batch size of 8. We also include Dropout \cite{dropout} in all our Lightweight Non-Bottleneck modules as regularizers. Following \cite{erfnet} we set the dropout rate of the first five modules to $0.03$ and all the others to $0.3$.

\subsection{Results on Cityscapes}
\label{sec:cityscapes_results}
Cityscapes \cite{cityscapes} is a large scale dataset for semantic urban scene understanding. It consists of 5000 finely annotated high-resolution images with pixel-level fine annotations. Images have been collected in 50 different cities around Europe, with high variability of weather conditions and in different seasons. We used the standard split suggested by the authors which consist in 2975, 500, and 1525 images for train, validation, and test sets respectively. Annotations include 19 classes used to train and evaluate models. Following a common practice for efficient oriented architectures \cite{erfnet,enet,espnet}, images have been subsampled by a factor 2 for every experiment reported on Cityscapes dataset.

\textbf{Improved GUM on efficient architectures} We investigated the use of iGUM module on GUNet \cite{gum} as a replacement for the original GUM module. Furthermore, we tested our iGUM within three state-of-the-art efficient architectures for semantic segmentation. We trained the networks with their original decoder and compared them with modified versions where the decoder is replaced by the improved GUM module. Table \ref{tab:gum_nogum} shows the results of these experiments: the first line shows the performance of GUM architecture with the original GUM and with our iGUM module. The results support the theoretical study exposed in Section \ref{sec:igum}: by replacing GUM with iGUM in GUNet architecture \cite{gum}, the loss in accuracy is negligible.  On the other hand, the model exhibits a visible benefit in speed. Both speed and mIoU improve over the baselines on all the other architectures by a large margin. The baseline does not correspond exactly with results reported by the papers because we trained these three networks with the same settings exposed in Section \ref{sec:experiments} which may differ from those used by the authors to train their own architecture. Moreover, they have not been pretrained and the speed has been evaluated by removing the Batch-Normalization layer. For the sake of these experiments, we are only interested in the relative performances.

\begin{table}[]
\begin{center}
\resizebox{0.6\linewidth}{!}{
\begin{tabular}{@{}lcllll@{}}
\toprule
Model & \multicolumn{2}{l}{Original Decoder} & \multicolumn{2}{l}{iGUM Module} \\ \midrule \midrule
       & mIoU              & FPS              & mIoU            & FPS           \\
GUNet \cite{gum} & \textbf{64.8} & 40 & 64.4 & \textbf{48} \\
ERFNet \cite{erfnet} & 60.7              & 59               & \textbf{63.7}            & \textbf{81}            \\
ENet \cite{enet}  & 47.3              & 87               & \textbf{55.7}            & \textbf{137}           \\
ESPNet \cite{espnet} & 48.2              & 172              & \textbf{52.9}            & \textbf{206}           \\ \bottomrule
\end{tabular}
}
\end{center}
\caption{mIoU and FPS on Cityscapes val set by  replacing the original decoder with our Improved Guidance Upsampling Module in GUNet and three other efficient architectures for semantic segmentation.}
\label{tab:gum_nogum}
\end{table}

\begin{table}
\begin{center}
\resizebox{0.9\linewidth}{!}{
\begin{tabular}{llllr}
\toprule
Encoder & Pretraining & Decoder & mIoU\% & FPS \\
\midrule \midrule
ERFNet (baseline) & \checkmark & ERFNet & \textbf{72.3} & 60.3 \\
ERFNet & \checkmark & improved GUM & 71.6 & 84.4 \\
Ours & \checkmark & improved GUM & 69.3 & \textbf{113.1} \\

\bottomrule
\end{tabular}
}
\end{center}
\caption{Ablation study on Cityscapes val dataset for our encoder and decoder. We adopted ERFNet \cite{erfnet} as baseline and replaced encoder and decoder in two steps. mIoU slighly decreases w.r.t. the baseline but the inference speed almost doubles. Encoders have been pre-trained on Imagenet.}
\label{tab:architecture_ablation}
\end{table}

\textbf{Ablation Studies} We designed our network starting from ERFNet \cite{erfnet} as baseline.
First, we replaced the decoder part, which in the original ERFNet is composed of three deconvolutions and four Non-Bottleneck-1D Modules. The introduction of the Improved GUM cause a negligible performance decrease, i.e. from 72.3 to 71.6 but improves speed by 24 frames per seconds. As a second step, we replaced also the encoder obtaining a completely new architecture. Again, a small decrease in performance but a large speedup compared to the baseline. Table \ref{tab:architecture_ablation} shows the results of the two ablation experiments.

\begin{table}
\begin{center}
\resizebox{0.9\linewidth}{!}{
\begin{tabular}{lllllllr}
Module Name & \begin{Turn} Conv 1x1 \end{Turn} & \begin{Turn} No Bias \end{Turn} & \begin{Turn} PReLU \end{Turn} & \begin{Turn} Moved 1x1 \end{Turn} & \begin{Turn} Pretrain \end{Turn} & mIoU\% & FPS\\
\midrule \midrule
Non-bt-1D (baseline) & & & & & & 63.2 & 84.4\\
 & \checkmark & & & & & 62.4 & 95.3 \\
 & & \checkmark & & & & 63.6 & 96.0 \\
 & & & \checkmark & & & 65.6 & 82.3 \\
 & \checkmark & \checkmark & \checkmark & & & 63.1 & \textbf{113.1} \\
Lightweight Non-bt-1D & \checkmark & \checkmark & \checkmark & \checkmark & & 64.1 & \textbf{113.1} \\ \midrule
Lightweight Non-bt-1D & \checkmark & \checkmark & \checkmark & \checkmark & \checkmark & \textbf{69.3} & \textbf{113.1} \\
\bottomrule
\end{tabular}
}
\end{center}
\caption{Ablation study for our Lightweight Non-Bottleneck 1D Module on Cityscapes val set.}
\label{tab:module_ablation}
\end{table}

In Table \ref{tab:module_ablation} we show the results of the ablation experiments for every choice made to obtain the Lightweight Non-Bottleneck-1D module. The baseline is an architecture composed by Non-bt-1D modules from \cite{erfnet}. A Non-bt-1D module is composed by four factorized 1D kernels. The first two have dilation term 1 whereas the last two have different dilation terms, i.e from 1 to 16, depending on the position within the architecture. We replaced the first two factorized convolutions with 1x1 convolutions. This speeded up the architecture by 15 FPS with a negligible mIoU decrease. Then we removed the biases from every convolution obtaining an mIoU of 63.6\% and a slight increase in speed. Inspired by \cite{enet} we replaced ReLUs with PReLUs. This increased mIoU by 2\% without any loss in speed. By applying these modifications together we got a mIoU of 63\% and a very fast encoder, i.e. 113.1 FPS. Finally we moved one 1x1 convolution to the end of the residual module like shown in Figure \ref{fig:lightweight_nbt_1d_module} obtaining an mIoU of 64.1\%. Last row shows the effect of pretraining on Imagenet.
\begin{table}[]
\begin{center}
  \resizebox{\linewidth}{!}{
\begin{tabular}{@{}llllll@{}}
\toprule
Method     & IoU class & iIoU class & IoU category & iIoU category & FPS   \\ \midrule
Mapillary \cite{mapillary}  & 82.0      & 65.9       & 91.2         & 81.7          & n/a   \\
PSPNet \cite{psp}    & 81.2      & 59.6       & 91.2         & 79.2          & 2.7   \\
FCN-8s \cite{fcn8s}     & 65.3      & 41.7       & 85.7         & 70.1          & 4.9   \\
DeepLabv3+ \cite{deeplabv3plus} & \textbf{82.1}      & \textbf{62.4}       & \textbf{92.0}         & \textbf{81.9}          & 5.1   \\
SegNet \cite{segnet}     & 57.0      & 32.0       & 79.1         & 61.9          & 20.1  \\
SQ \cite{sq}        & 59.8      & 32.3       & 84.3         & 66.0          & 28.7  \\
GUNet \cite{gum}      & 70.4      & 40.8       & 86.8         & 69.1          & 37.0  \\
ERFNet \cite{erfnet}     & 69.7      & 44.1       & 87.3         & 72.7          & 61.0  \\
ContextNet \cite{contextnet} & 66.1      & 36.8       & 82.8         & 64.3          & 62.0  \\
ENet \cite{enet}      & 58.3      & 34.4       & 80.4         & 64.0          & 88.1  \\
ESPNet \cite{espnet}     & 60.3      & 31.8       & 82.2         & 63.1          & 112.0 \\
Ours       & 68.9      & 39.0       & 85.9         & 66.5          & \textbf{113.1} \\ \bottomrule
\end{tabular}
}
\end{center}
\caption{Comparison with representative architectures for semantic segmentation on Cityscapes test set. mIoU evaluated by Cityscapes evaluation server. FPS reported from original paper if authors used a TitanX (Pascal) GPU, otherwise FPS computed on our GPU.}
\label{tab:cityscapes_comparison}
\end{table}

\label{sec:instance_cityscapes}
\begin{table}[]
\begin{center}
\resizebox{0.45\linewidth}{!}{
\begin{tabular}{@{}llll@{}}
\toprule
Loss Function & mIoU & AP   & AP50\% \\ \midrule
L2            & 65.7 & \textbf{10.7} & \textbf{20.2}   \\
L1            & 65.3 & 9.8  & 18.0   \\
Smooth L1     & \textbf{66.7} & 10.2 & 18.3   \\ \bottomrule
\end{tabular}
}
\end{center}
\caption{Different loss functions to train our instance segmentation module. Results on Cityscapes val set.}
\label{tab:instance_loss}
\end{table}

\begin{table}[]
\begin{center}
\resizebox{0.5\linewidth}{!}{
\begin{tabular}{@{}ccccc@{}}
\toprule
Iterations & AP   & AP50\% & ms  & FPS   \\ \midrule
40         & 10.7 & 20.2   & 10.0 & 105.1 \\
30         & \textbf{10.7} & \textbf{20.2}   & 9.0 & 106.4 \\
20         & 10.5 & 20.0   & 8.9 & 107.4 \\
15         & 10.3 & 19.3   & 8.8 & 109.3 \\
10         & 9.7  & 17.7   & 8.8 & 109.7 \\
7          & 8.5  & 16.0   & 8.7 & 109.7 \\
5          & 7.2  & 14.3   & 8.7 & 110.1 \\
3          & 4.0  & 8.6    & \textbf{8.7} & \textbf{110.2} \\ \bottomrule
\end{tabular}
}
\end{center}
\caption{Experiments to assess the impact on performance and speed w.r.t the number of iterations of sampling in the Instance Segmentation Module. Tested on Cityscapes val set.}
\label{tab:iterations}
\end{table}

\begin{table}[]
\begin{center}
\resizebox{\linewidth}{!}{
\begin{tabular}{@{}lccccc@{}}
\toprule
name                         & AP   & AP 50\% & AP 100m & AP 50m & FPS \\ \midrule
Deep Contours \cite{deepcontours}                & 2.3  & 3.7     & 3.9     & 4.9    & 5.0             \\
R-CNN + MCG convex hull \cite{cityscapes}     & 4.6  & 12.9    & 7.7     & 10.3   & 0.1              \\
FCN+Depth \cite{fcndepth}                    & 8.9  & 21.1    & 15.3    & 16.7   & n/a             \\
Joint Graph Decomposition \cite{levinkov2017joint}   & 9.8  & 23.2    & 16.8    & 20.3   & n/a             \\
Boundary-aware \cite{hayder2017boundary}              & 17.4 & 36.7    & 29.3    & 34     & n/a             \\
Discriminative Loss Function \cite{discriminativeloss} & 17.5 & 35.9    & 27.8    & 31     & n/a             \\
Dist. Watershed Transform \cite{bai2017deep}   & 19.4 & 35.3    & 31.4    & 36.8   & n/a             \\
Fast Scene Understanding \cite{neven2017fast}     & 21.0 & 38.6    & 34.8    & 38.7   & 21.3           \\
Multitask Learning \cite{kendall2017multi}          & 21.6 & 39      & 35      & 37     & n/a             \\
Mask R-CNN \cite{maskrcnn}                  & 26.2 & 49.9    & 37.6    & 40.1   & n/a             \\
PANet \cite{panet}                       & \textbf{31.8} & \textbf{57.1}    & \textbf{44.2}    & \textbf{46}     & n/a             \\
Ours                         & 9.2  & 16.8    & 16.4    & 21.4   & \textbf{106.4}           \\ \bottomrule
\end{tabular}
}
\end{center}
\caption{Comparison with State of the art methods on Cityscapes test set from the cityscapes leaderboard.}
\label{tab:instance_sota}
\end{table}
\textbf{Instance Segmentation}
With the first cluster of experiments we want to determine the best loss function to train our architecture for the instance segmentation task. Table \ref{tab:instance_loss} shows results on the Cityscapes validation set in terms of mIoU on semantic segmentation and AP on instance segmentation. Surprisingly the variance is low between different loss functions. We decided to keep $L^2$ loss for the next experiments.
With our second cluster of experiments we want to assess how many sampling iterations $t$ (see Section \ref{sec:instance}) are needed for convergence and how much they affect performance and speed. Table \ref{tab:iterations} shows the results for these experiments. The FPS are not very affected, due to the efficiency of the sampling module, the performances start to degrade significantly from 15 iterations. We decided to keep 30 iterations for the next experiment. We tested the network against the Cityscapes test set: results are shown in Table \ref{tab:instance_sota}. Our method achieve the 9.2\% of AP which is far from the state-of-the-art accuracy-oriented methods like PANet \cite{panet} but it exhibit a remarkably fast inference time.
\subsection{Results on Camvid}
\begin{table}
\begin{center}
\resizebox{0.8\linewidth}{!}{
\begin{tabular}{lcccc}
\toprule
Method & Pretraining & Class avg. & mIoU & Global avg. \\ \midrule \midrule
Segnet \cite{segnet}               & \checkmark & 65.2 & 55.6 & 88.5 \\
ENet \cite{enet}                  &       & 68.3 & 51.3 & n/a  \\
ESPNet \cite{espnet}               &       & 68.3 & 55.6 & n/a  \\
ERFNet \cite{erfnet} &       & 65.8 & 53.1 & 86.3 \\
ERFNet \cite{erfnet} & \checkmark & 72.5 & 62.7 & 89.4 \\
FCN-8s \cite{fcn8s}               & \checkmark & n/a  & 57.0 & 88.0 \\
Dilation8 \cite{drn}           & \checkmark & n/a  & 65.3 & 79.0 \\
DeepLab \cite{deeplab}             & \checkmark & n/a  & 61.6 & n/a  \\
Ours     & \checkmark & \textbf{76.9} & \textbf{68.7} & \textbf{91.9} \\ \bottomrule
\end{tabular}
}
\end{center}
\caption{Results on Camvid test set ordered by increasing mIoU. Our model outperform every other efficient architecture by a large margin. It even yelds better results with respect to some accuracy-oriented architectures.}
\label{tab:camvid}
\end{table}
We tested our architecture on the Camvid\cite{camvid} dataset. It is composed of 367 training and 233 testing images of urban outdoor environments. It has been tagged in eleven semantic classes of which one is not evaluated and thus not used for training. The original frame resolution is $970\times720$. Following \cite{enet,segnet,espnet}, for fair comparison, we downsampled the images to $480\times360$ pixels before training. This dataset represents an interesting benchmark to test the behaviour of our method with low-cardinality datasets. In Table \ref{tab:camvid} we compare the performance of our architecture with existing state-of-the-art efficient architectures. We also include three computationally-heavy architectures. Our architecture yields very good results with respect to other efficiency-oriented architectures and even outperforms some accuracy-oriented methods.
\subsection{Results on PASCAL VOC 2012}

\begin{table}
\begin{center}
\resizebox{0.5\linewidth}{!}{
\begin{tabular}{p{4cm} c}
\toprule
Method & mIoU\% \\ \midrule \midrule
SegNet \cite{segnet}    & 59.10 \\
LRR \cite{lrr}       & 79.30 \\
Dilation-8 \cite{drn} & 75.30 \\
FCN-8s \cite{fcn8s}    & 67.20 \\
ESPNet \cite{espnet}    & 63.01 \\
DeepLab \cite{deeplab}   & 79.70 \\
RefineNet \cite{refinenet} & 82.40 \\
PSPNet \cite{psp}    & 85.40 \\
DeepLabv3+ \cite{deeplabv3plus} & \textbf{87.80} \\
Ours       & 63.54 \\ \bottomrule
\end{tabular}
}
\end{center}
\caption{Results on PASCAL VOC 2012 test set. We reported some popular methods and the state-of-the-art on Pascal VOC.}
\label{tab:pascalvoc}
\end{table}

We tested our network architecture against the popular PASCAL VOC 2012 segmentation dataset \cite{pascalvoc} which contains 20 object categories and a background class. It is composed of 1464, 1448 and 1456 images for the training, validation and test sets respectively. Following \cite{fcn8s, chen2014semantic, papandreou2015weakly, dai2015boxsup} we used additional images to train our network with data annotation of \cite{hariharan2011semantic} resulting in 10582 1449 and 1456 images for training, validation and testing. Table \ref{tab:pascalvoc} shows a comparison of our method with state-of-the-art popular architectures. Only ESPNet and our architecture are speed-oriented while all the others focus on accuracy. Notice that, besides being computationally heavy PSPNet and DeepLabV3+ have been pretrained on COCO \cite{lin2014microsoft} dataset.
\subsection{Speed}
\begin{table}[]
\begin{center}
\resizebox{\linewidth}{!}{
\begin{tabular}{@{}cccccccccccc@{}}
\toprule
\multicolumn{12}{c}{NVIDIA TEGRA TX1 (Jetson)}                                                                                                                                          \\
ms          & fps           & ms           & fps           & ms            & fps           & ms          & fps           & ms           & fps           & ms            & fps           \\
\multicolumn{2}{c}{$640\times360$} & \multicolumn{2}{c}{$1280\times720$} & \multicolumn{2}{c}{$1920\times1080$} & \multicolumn{2}{c}{$512\times256$} & \multicolumn{2}{c}{$1024\times512$} & \multicolumn{2}{c}{$2048\times1024$} \\ \midrule
109         & 9.9           & 471          & 2.1           & 1.43          & 0.7           & 49          & 20.5          & 207          & 5.1           & 1228          & 0.9           \\ \midrule
\multicolumn{12}{c}{NVIDIA TITAN Xp}                                                                                                                                                    \\
\multicolumn{2}{c}{$640\times360$} & \multicolumn{2}{c}{$1280\times720$} & \multicolumn{2}{c}{$1920\times1080$} & \multicolumn{2}{c}{$512\times256$} & \multicolumn{2}{c}{$1024\times512$} & \multicolumn{2}{c}{$2048\times1024$} \\ \midrule
4           & 235.9         & 13           & 79.1          & 29            & 34.4          & 4           & 242.1         & 8            & 113.1         & 29            & 34.7          \\ \bottomrule
\end{tabular}
}
\end{center}
\caption{Network speed on a high-end GPU and an embedded device.}
\label{tab:speed}
\end{table}

\begin{figure}
\centering
\includegraphics[width=0.9\linewidth]{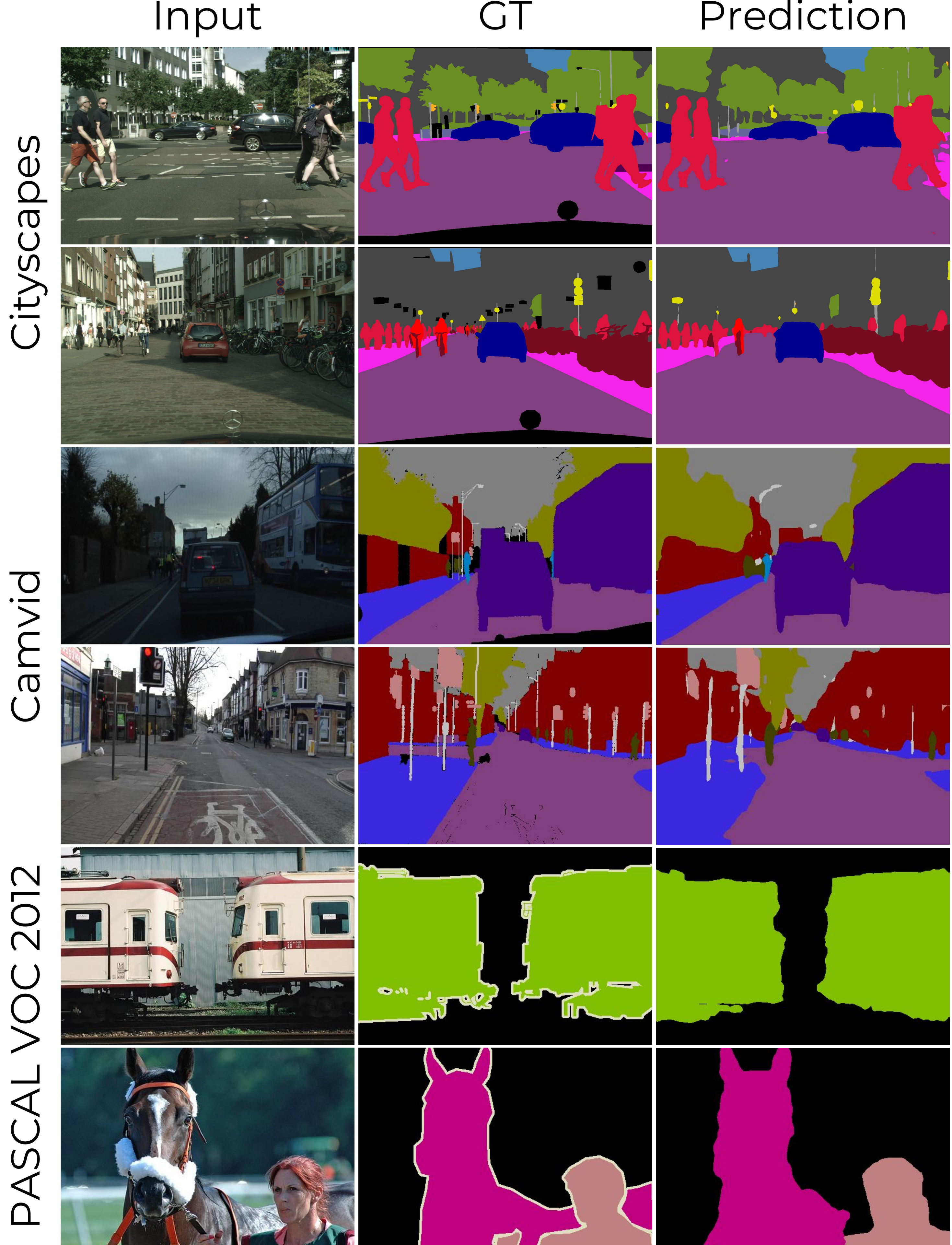}
\caption{Visual results on images from three datasets: Cityscapes, Camvid and PASCAL VOC 2012}
\label{fig:visual_results}
\end{figure}

We report in Table \ref{tab:speed} the speed of our network on a high-end Titan Xp GPU with Pytorch 1.0 and CUDA 10.0. Then we tested our architecture on an edge device: Nvidia Tegra TX 1 (Jetson) with Pytorch 0.4.

\section{Concluding Remarks}
We presented a novel architecture for efficient scene understanding which includes a novel module to speed up the decoder part of encoder-decoder architectures and a module for Instance Segmentation based on iterative sampling. We tested our architecture on three different datasets showing that our network is fast and accurate compared to state-of-the-art efficient architectures.

{\small \textbf{Acknowledgments} This work was supported by TEINVEIN, CUP: E96D17000110009 - Call "Accordi per la Ricerca e l’Innovazione", cofunded by POR FESR 2014-2020.
We gratefully acknowledge the support of NVIDIA Corporation with the donation of a Titan Xp GPU used for this research.}

{\small
\bibliographystyle{ieee}
\bibliography{paper}
}

\end{document}